\documentclass{article}
\usepackage{spconf,amsmath,graphicx,color, tabularx}
\usepackage{booktabs, multirow, color}
\usepackage{float}

\title{Can Generative Large Language Models perform ASR error correction?}

\name{Rao Ma, Mengjie Qian, Potsawee Manakul, Mark Gales, Kate Knill\thanks{This paper reports on research supported by Cambridge University Press \& Assessment, a department of The Chancellor, Masters, and Scholars of the University of Cambridge. Mengjie Qian is supported by EPSRC Project EP/V006223/1 (Multimodal Video Search by Examples).}}
\address{ALTA Institute, Machine Intelligence Lab, Department of Engineering, Cambridge University, UK}
\begin{document}
%
\maketitle
\begin{abstract}
ASR error correction is an interesting option for post processing speech recognition system outputs. These error correction models are usually trained in a supervised fashion using the decoding results of a target ASR system. This approach can be computationally intensive and the model is tuned to a specific ASR system. Recently generative large language models (LLMs) have been applied to a wide range of natural language processing tasks, as they can operate in a zero-shot or few shot fashion. In this paper we investigate using ChatGPT, a generative LLM, for ASR error correction. Based on the ASR N-best output, we propose both unconstrained and constrained, where a member of the N-best list is selected, approaches. Additionally, zero and 1-shot settings are evaluated. Experiments show that this generative LLM approach can yield performance gains for two different state-of-the-art ASR architectures,  transducer and attention-encoder-decoder based, and multiple test sets.

\end{abstract}
\section{Introduction}
\label{sec:intro}
Automatic speech recognition (ASR) systems aim to transcribe human speech into readable text and are the key component for human-computer interaction \cite{rebman2003speech}. In recent years, significant advancements have been made in this area. End-to-end (E2E) systems such as LAS  or RNN-T are effective at modelling long contexts within the utterance and show superior performance compared to the HMM-based counterparts \cite{chan2016listen, graves2014towards, amodei2016deep}. The training of ASR systems requires the availability of high-quality transcribed speech data, which can be costly to obtain.
Publicly available corpora usually contain at most thousands of hours of annotated speech data. In contrast, the recently released ASR model, 
Whisper~\cite{radford2023robust}, is pre-trained on around 680,000 hours of weakly supervised data collected from the Internet. 

The decoder part of an RNN-T or a LAS model acts as a language model that estimates the probability of the generated word sequence \cite{meng2021internal}. It learns from the reference text and is jointly trained with the acoustic encoder. Due to the limited availability of labelled speech training data, ASR systems struggle to generate rare words that have low frequency in the training corpus.
Large quantities of text data covering a wide range of domains are much easier to collect and process than speech data. Therefore, text-based methods have been explored to improve the performance of speech recognition systems. 
One option is ASR error correction which automatically identifies errors within the ASR hypothesis and outputs the corrected transcription~\cite{errattahi2018automatic, leng2021fastcorrect, wang2020asr}. 

The development of the error correction model follows the trend of Natural Language Processing (NLP) technology. Early models were rule-based systems, which required carefully designed features and human expertise \cite{cucu2013statistical}. With the emergence of recurrent networks and attention mechanisms, models with an E2E architecture became mainstream. 
These models usually adopt a similar structure where the bidirectional encoder takes the ASR transcription as input and the reference text is used as the training target.
This approach has shown promising performance on diverse datasets for ASR models of different architectures~\cite{guo2019spelling, hrinchuk2020correction, ma2020neural, zhu2021improving}.

In the past few years, large-scale pre-trained language models became available. These are generally trained on multi-domain text data that is several magnitudes more than the prevailing ASR systems. For instance, BERT is pre-trained on 3,300M words~\cite{devlin2019bert} and T5 is trained on 750GB text~\cite{raffel2020exploring}.
Previous works~\cite{ma2023nbest,ma2023adapting} developed methods to build an ASR error correction model based on the powerful T5 model. By fine-tuning from the pre-trained NLP model, implicit knowledge learned from huge amounts of text data can be effectively transferred to the target error correction task. 
Results indicate the importance of adopting the ASR N-best list rather than the top one hypothesis as model input for accessing richer context in the correction process.

Typically error correction models are trained in a supervised fashion to effectively learn the error patterns made by the ASR system. The training process requires first decoding large amounts of speech data with the ASR system of interest, and then using the erroneous hypotheses to train the correction model. These two stages can be computationally intensive to adopt in practice. Additionally, the error correction model is usually bound to a specific ASR system and a particular domain. Therefore, when we switch the underlying ASR system or apply it to a new domain, the corresponding error correction model can be less effective and needs to be re-trained. To address the above issues, in this paper we propose approaches to perform zero-shot or few-shot ASR error correction. These novel methods are training-free and enable plug-and-play support to an existing ASR system.

Generative large language models (LLMs) such as ChatGPT have demonstrated remarkable performance of language understanding on text processing tasks \cite{openai2023gpt4, touvron2023llama, anil2023palm}. In our work, we examine its performance in identifying and correcting errors on two state-of-the-art ASR architectures. Different prompts to ChatGPT and both unconstrained and constrained generation methods are compared on three standard ASR data sets. Zero and 1-shot settings are evaluated as well. The results show that this efficient generative LLM, ChatGPT, error correction approach yields performance gains for both ASR architectures.






\section{Background}
\label{sec:nbest_t5}
Error correction models aim to fix errors in the ASR transcription and are an interesting option for ASR post-processing.
A standard error correction model adopts an E2E structure, taking the ASR transcriptions as the model input and generating the corrected sentence. 
Several model variants incorporating additional inputs have been proposed \cite{leng2021fastcorrect, wang2020asr, ma2020neural}.
\cite{ma2023nbest} proposes an N-best T5 error correction model that is fine-tuned from a pre-trained T5 model. It leverages the ASR N-best  hypotheses as model input and demonstrates significant performance gain over the model using the 1-best input. It also proposes an N-best constrained decoding approach in error correction, which uses the combined scores of the ASR model and the T5 model to find the best hypothesis in the N-best list. 
\begin{figure}[!htbp]
    \centering
    \includegraphics[width=0.97\linewidth]{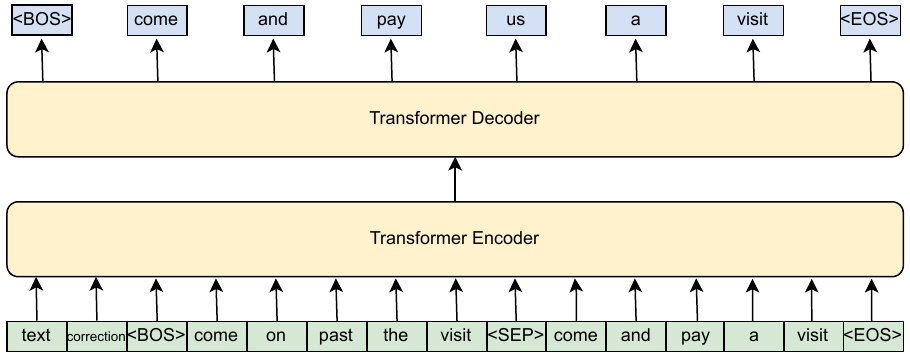}
    \caption{N-best T5 error correction model structure.}
    \label{fig:error_correction}
\end{figure}


There has been rapid growth in the LLM literature, and larger and better LLMs are constantly being released. 
LLMs continue to be scaled up in size and pre-trained on increasingly more data. 
With reinforcement learning from human feedback (RLFH), they are capable of performing several NLP tasks in a zero-shot manner~\cite{instructgpt2022ouyang, liu2023summary}. 
For example, LLMs such as ChatGPT have been applied to summary assessment~\cite{luo2023chatgpt}, and grammatical error correction~\cite{wu2023chatgpt, fang2023chatgpt}. Their inherent ability to perform ASR post-processing tasks, however, has been less explored. In this work, we follow \cite{ma2023nbest} to use the ASR N-best list as input to the error correction model while using the more powerful ChatGPT model rather than T5 to perform the task. 

\section{LLM-Based ASR Error Correction}
\begin{figure*}[t]
    \centering
    \includegraphics[trim=0 1mm 0 5mm, width=0.83\linewidth]{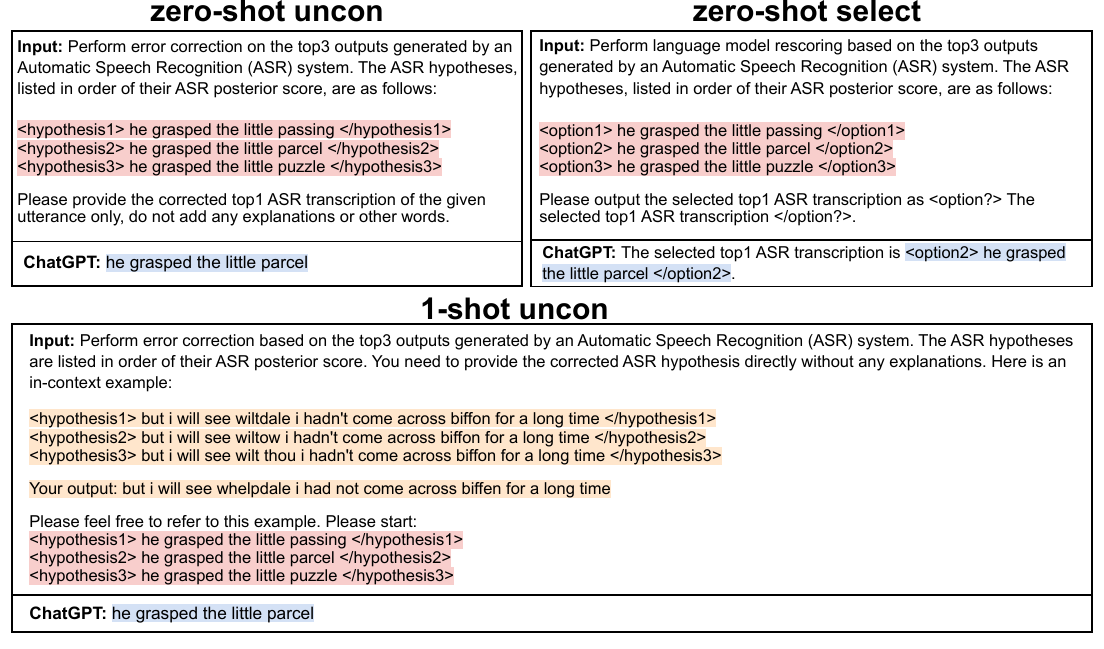}
    \vspace{-0.3cm}
    \caption {Prompt design for (a) zero-shot unconstrained error correction, (b) zero-shot selective approach, and (c) 1-shot unconstrained error correction. Here we use a 3-best list generated by the ASR system as input to ChatGPT for illustration.}
    \label{fig:prompt}
    \vspace{-0.2cm}
\end{figure*}

In this section, we introduce our methods of utilising generative large language models for zero-shot or few-shot error correction. Two types of tasks are discussed: unconstrained error correction and N-best constrained error correction.

\subsection{Unconstrained Error Correction}
In the unconstrained error correction \textbf{(uncon)} setting, we ask ChatGPT to directly output the corrected hypothesis without adding an explanation. This task can be relatively difficult to perform as ChatGPT has no prior knowledge about the error patterns of the ASR system and no access to the original utterance. Instead of the 1-best ASR transcription, therefore, we input the N-best list obtained from the beam search decoding of the ASR model to ChatGPT. Hypotheses from the N-best list can act as hints to help the model better detect and correct the errors \cite{liu2021asr, ganesan2021n}. In the ablation study in Section~\ref{sec:exp-ablation} we show that using a reasonable number of N is important for the model to achieve good performance. 
When only the top one ASR hypothesis is used as input ChatGPT-based error correction may degrade performance.

\subsubsection{Zero-shot vs. 1-shot Prompts}
The prompt designed for the \textit{zero-shot uncon} setting is illustrated in Figure \ref{fig:prompt}. In this prompt the hypotheses are sorted by the (descending) ASR posterior score. Furthermore, tags like \texttt{$<$hypothesis1$>$} and \texttt{$<$/hypothesis1$>$} are used to surround each N-best hypothesis. Other input formats such as using numbers rather than tags or using plain sentences without the explicitly specified order were also examined and showed degraded performance to our selected prompt. 
Considering the complexity of this task, we additionally experiment with the 1-shot setting to perform in-context learning.
Here, we give an example for ChatGPT to refer to before conducting error correction (highlighted in orange in Figure \ref{fig:prompt}). This example is selected from a decoding result of the Transducer ASR model on the dev\_other set of LibriSpeech. It is used in all our 1-shot error correction experiments. 
By showing both input and the desired output in the prompt, we hope to remind ChatGPT to match the sentence length of the given hypotheses and only make edits to the detected errors.

\subsection{N-best Constrained Error Correction}
Unconstrained error correction is the standard approach to generate the corrected transcription based on the information from the given hypotheses. Results in \cite{ma2023nbest, ma2023adapting} suggest that constraining the decoding space to the given N-best list leads to performance gains in some cases. We propose two methods, therefore, to constrain the output of ChatGPT to be a hypothesis within the given N-best list, namely the selective approach and the closest mapping.

\subsubsection{Selective Approach}
With the selective approach \textbf{(select)}, ChatGPT is asked to select the most likely ASR transcription from all the candidates rather than generate one from scratch. All the input sentences are listed as \texttt{$<$option1$>$ ASR hypothesis $<$/option1$>$}, and ChatGPT is asked to return the selected option in the format of \texttt{<option?> The selected ASR transcription </option?>}. This method is similar to language model rescoring to some extent, however, it performs the selection in one go. More importantly, ChatGPT sees all the candidates before deciding on the best one. This is different from the rescoring process where language model scores are generated individually for each of the N-best hypotheses without comparing the similarity and correlation between each other.

\subsubsection{Closest Mapping}
The closest mapping method \textbf{(closest)} is based on the assumption that when ChatGPT performs unconstrained error correction, it first selects the best hypothesis from the given N-best list and then makes modifications based on this sentence to yield the final output. We hope to find this \textit{``closest match''} in a reverse process by finding the hypothesis within the ASR N-best list that has the smallest Levenshtein distance to the ChatGPT unconstrained generation result. For instance, for the \textit{zero-shot uncon} example in Figure \ref{fig:prompt} the Levenshtein distance of the ChatGPT output to the 3-best ASR hypotheses is $1, 0, 1$ respectively. The second hypothesis will be selected, therefore, as the corrected result for this utterance.

\section{Experiments}
\subsection{Setup}







We conduct experiments on ChatGPT (gpt-3.5-turbo-0613) to study its performance on error correction for two ASR models. 
One is a Conformer-Transducer \cite{gulati2020conformer} model containing 12 encoder layers. The model was trained on 960 hours LibriSpeech data with SpecAugment~\cite{park2019specaugment} and speed perturbation applied, following the ESPnet recipe~\cite{watanabe2018espnet}. The other ASR model studied is the Whisper~\cite{radford2023robust} small.en model. 
In decoding we suppress the probability of the most common punctuation marks as in \cite{ma2023adapting}.
Each ASR model is decoded with a beam size of 10 that generates a 10-best list as a by-product at inference. If not stated otherwise, the top five hypotheses are used as input to ChatGPT, i.e. the size of the input N-best list is 5. The effect of adopting different N is studied in the ablation experiment. 
We lowercase the ASR N-best list without performing other text processing steps prior to input to ChatGPT.
At the evaluation stage, we run the text normalisation scripts from the Whisper project on both the ASR reference and the hypothesis text before calculating WER results.

\begin{table}[!htbp]
    \centering
    \begin{tabular}{l|c|c|c|c}
    \toprule
        Dataset & Subset & \# Utts & \# Words & Hours \\ 
        \midrule
        LibriSpeech & test\_other & 2,939 & 52K & 5.3 \\ 
        TED-LIUM3 & test & 1,155 & 52K & 4.6 \\ 
        Artie Bias & test & 1,712 & 15K & 2.4\\ 
    \bottomrule
    \end{tabular}
    \caption{Statistics of test sets used in the experiments.}
    \label{tab:statics}
    \vspace{-0.4cm}
\end{table}

The proposed approaches are evaluated on three public datasets, namely LibriSpeech~\cite{panayotov2015librispeech}, TED-LIUM3~\cite{hernandez2018ted}, and the Artie Bias Corpus~\cite{meyer2020artie}. LibriSpeech is an audiobook-based English speech corpus, TED-LIUM3 is an audio dataset collected from TED talks, and the Artie Bias Corpus is a subset of the Common Voice dataset~\cite{ardila2020common} which is also read speech. The details of the datasets are presented in Table~\ref{tab:statics}. We undertook a comparative analysis between ASR error correction using the generative LLM ChatGPT and a standard error correction model that adopts an E2E structure. Specifically, we trained N-best T5 error correction models as described in Section~\ref{sec:nbest_t5}. An error correction model was trained for each of the two ASR systems mentioned above. Each N-best T5 model was fine-tuned on the 10-best outputs of the associated ASR model decoded on the 960 hours LibriSpeech training set.





\subsection{Experimental Results}

In Table~\ref{tab:ct_correction}, we study the behaviour of ChatGPT on ASR error correction when a Transducer-based model or a pre-trained Whisper model with an Attention-Encoder-Decoder (AED) structure is used as the base ASR system. Results from the fine-tuned N-best 
T5 error correction model are listed for comparison. The best performance of the T5 model and ChatGPT based experiments are highlighted in bold. 

For the Conformer-Transducer model, 
LibriSpeech can be considered as an in-domain dataset. In this case, the supervised trained T5 model yields a relative gain of 10.9\% (6.90\% to 6.15\%) over the ASR baseline. 
Unlike the T5 model, ChatGPT does not require any form of model training prior to error correction and is therefore more efficient.
In the zero-shot setting, both the selective approach and closest mapping perform better than the unconstrained generation.
The \textit{0-shot closest} which finds the closest match of the output corrected hypothesis in the given N-best list performs better than asking ChatGPT to directly select the best one from the N-best list. The unconstrained error correction results become much better when we switch to the \textit{1-shot uncon} prompt (6.64\% to 6.29\%), indicating that ChatGPT better understands the task by referring to the given example. When we apply the closest mapping in the 1-shot setting, the test set WER is reduced to 6.24\%, which is comparable to the T5 model performance.

\begin{table}[tbp!]
    \centering
    \begin{tabular}{@{}l@{ }|p{6mm}p{6mm}|p{7.3mm}p{7.3mm}|p{7.5mm}p{7.5mm}@{ }}
        \toprule
        \multirow{2}*{System} & \multicolumn{2}{c|}{LB} & \multicolumn{2}{c|}{TED} & \multicolumn{2}{c}{Artie} \\
        & CTr & Whs & CTr & Whs & CTr & Whs \\
        \midrule
        Baseline & 6.90 & 7.37 & 13.53 & 3.89 & 23.67 & 9.03 \\
        \midrule
        Oracle & 4.59 & 5.24 & 10.71 & 2.59 & 17.95 & 5.59 \\
        \midrule
        \multirow{1}*{T5} (uncon) &  6.37$^\dagger$ & \textbf{6.39}$^\dagger$ & \textbf{12.00}$^\dagger$ & 4.56 & \textbf{21.24}$^\dagger$ & 9.16 \\
        \multirow{1}*{T5} (constr) & \textbf{6.15}$^\dagger$ & 6.69$^\dagger$  & 12.12$^\dagger$ & \textbf{3.64}$^\dagger$ & 21.36$^\dagger$ & \textbf{8.14}$^\dagger$ \\
        \midrule\midrule
        \multicolumn{7}{c}{\bf ChatGPT Error Correction}\\
        \midrule
        0-shot uncon & 6.64 & 7.71 & 11.35$^\dagger$ & 5.84 & \textbf{18.73}$^\dagger$ & 8.30$^\dagger$ \\
        0-shot select &  6.52$^\dagger$ & 7.24 & 12.61$^\dagger$ & \textbf{4.19} & 21.88$^\dagger$ & 8.47$^\dagger$ \\
        0-shot closest &  6.29$^\dagger$ & 7.15 & 11.97$^\dagger$ & 4.56 & 20.64$^\dagger$ & \textbf{8.21}$^\dagger$ \\
        \midrule
        1-shot uncon & 6.29$^\dagger$ & 7.18 & \textbf{10.13}$^\dagger$ & 4.96 & 19.35$^\dagger$ & 8.45$^\dagger$ \\
        1-shot closest & \textbf{6.24}$^\dagger$ & \textbf{7.03}$^\dagger$ & 11.96$^\dagger$ & 4.58 & 21.13$^\dagger$ & 8.53\\
        \bottomrule
    \end{tabular}
    \caption{WER results for a Conformer-Transducer (CTr) system and Whisper (Whs). Error correction results using a T5 model and ChatGPT are compared. $\dagger$ indicates the improvement over baseline is statistically significant with $p < 0.001$.}
    \label{tab:ct_correction}
    \vspace{-0.4cm}
\end{table}

When error correction is applied to Whisper on LibriSpeech, the unconstrained T5 model yields a 13.3\% WERR (7.37\% to 6.39\%). Less improvement is seen with the ChatGPT approach. Although the \textit{0-shot uncon} method performs worse than the ASR baseline, with N-best constrained error correction such as closest mapping, ChatGPT improves the performance over the original Whisper output. The best performance is also obtained with the \textit{1-shot closest} setting, achieving a WER of 7.03\% on the LibriSpeech test set.

For the Conformer-Transducer model, TED-LIUM3 and Artie Bias can be considered as out-of-domain datasets. The ASR system shows high error rates on these test sets while the T5-based error correction gives 11.3\% and 10.2\% WERR. Results from the ChatGPT-based methods show significant performance improvement compared to the T5 model. On the TED-LIUM3 test set, the \textit{1-shot uncon} approach outperforms the ASR baseline by 25.1\%. The result is even better than the oracle WER of the 5-best list output by the ASR model. 
ChatGPT leads to worse performance, however, when correcting the Whisper model outputs. In particular, many more deletion errors than in the baseline can be observed in the ChatGPT outputs from all the proposed methods.

On Artie Bias, \textit{0-shot uncon} yields the best performance for the Transducer ASR system, with 20.9\% WERR (23.67\% to 18.73\%) over the baseline result. \textit{1-shot uncon} performs slightly worse than the zero-shot setting. Since we pick the example in the 1-shot prompt from the dev set of LibriSpeech, this might lead to a mismatch on the Artie test set. Although Whisper already demonstrates state-of-the-art performance, both T5 and ChatGPT-based error correction approaches yield gains of 9.8\% and 9.0\% WERR, respectively.




\begin{table*}[h]
    \centering
    \small
    \begin{tabular}{l|l|l}
\toprule
ASR Model & Type & Text \\
\midrule
\multicolumn{2}{l|}{Reference Text} & now that that blew my mind and you know it took a lot of preparation we had to build cameras and lights ... \\
\midrule
\multirow{6}*{Transducer}&Hyp-1 & now that that {\bf\color{red}blue} my mind and you know {\bf\color{red}i} took a lot of preparation we had to build cameras and lights ... \\
&Hyp-2 & now that that {\bf\color{red}blue} my mind and you know {\bf\color{red}i} took a lot of preparation we had to build {\bf\color{red}camers} and lights ... \\
&Hyp-3 & now that that {\bf\color{blue}blew} my mind and you know {\bf\color{red}i} took a lot of preparation we had to build cameras and lights ... \\
&Hyp-4 & now that that {\bf\color{red}blow} my mind and you know {\bf\color{red}i} took a lot of preparation we had to build cameras and lights ... \\
&Hyp-5 & now that that {\bf\color{blue}blew} my mind and you know {\bf\color{red}i} took a lot of preparation we had to build {\bf\color{red}camers} and lights ...  \\
\cmidrule{2-3}
&ChatGPT & now that that {\bf\color{blue}blew} my mind and you know {\bf\color{red}i} took a lot of preparation we had to build cameras and lights ... \\
\midrule
\multirow{6}*{Whisper}&Hyp-1 & now that {\bf\color{red} ****} blew my mind and you know it took a lot of preparation we had to build cameras and lights ... \\
&Hyp-2 & now that {\bf\color{red} ****} blew my mind and {\bf\color{red} *** ****} it took a lot of preparation we had to build cameras and lights ... \\
&Hyp-3 & now that {\bf\color{red} ****} blew my mind and you know it took a lot of preparation we had to build cameras and lights ... \\
&Hyp-4 & now that {\bf\color{red} ****} blew my mind and {\bf\color{red} *** ****} it took a lot of preparation we had to build cameras and lights ... \\
&Hyp-5 & now that {\bf\color{red} ****} blew my mind and {\bf\color{red} *** ****} it took a lot of preparation {\bf\color{red}and} we had to build cameras and lights ... \\
\cmidrule{2-3}
&ChatGPT & now that {\bf\color{red} ****} blew my mind and {\bf\color{red} *** ****} it took a lot of preparation {\bf\color{red}and} we had to build cameras and lights ...\\
\bottomrule
    \end{tabular}
    \caption{Case analysis for ChatGPT \textit{1-shot uncon} error correction results with Conformer-Transducer and Whisper N-best lists.}
    \label{tab:case_analysis}
    \vspace{-0.2cm}
\end{table*}


\subsection{Discussion}
In Table \ref{tab:ct_lb_wer}, we calculate the WER breakdown of different types of errors for both ASR models. For the Conformer-Transducer model, when using the \textit{zero-shot uncon} prompt, the error correction results from ChatGPT contain fewer substitution and insertion errors compared to the original ASR baseline while causing much more deletions. With human evaluation, we find out that in the ChatGPT output, error correction results for 14 sentences are truncated (only the first few words are in the ChatGPT output rather than the entire sentence), contributing to 0.2\% absolute WER. With 1-shot learning, the ChatGPT output is more stable and all the problem cases are solved, yielding better overall performance.


\newcommand{\STAB}[1]{\begin{tabular}{@{}c@{}}#1\end{tabular}}

\begin{table}[!htbp]
    \centering
    \begin{tabular}{c|l|c|c|c|c}
    \toprule
       & \multirow{2}*{Method} & \multicolumn{4}{c}{WER} \\
       & & All & Sub & Del & Ins \\
       \midrule\midrule
       \multirow{4}{*}{\STAB{\rotatebox[origin=c]{90}{\bf Transducer}}}  & ASR baseline & 6.9 & 5.3 & 0.7 & 0.8  \\
        \cmidrule{2-6}
        & ChatGPT (0-shot uncon) & 6.6 & 4.6 & 1.3 & 0.7  \\
        & ChatGPT (1-shot uncon) & 6.3 & 4.6 & 0.9 & 0.8 \\
        & ChatGPT (1-shot closest) & 6.2 & 4.8 & 0.7 & 0.8 \\
        \midrule\midrule
       \multirow{4}{*}{\STAB{\rotatebox[origin=c]{90}{\bf Whisper}}} & ASR baseline & 7.4 & 4.9 & 1.7 & 0.8 \\
    \cmidrule{2-6}
        & ChatGPT (0-shot uncon)  & 7.7 & 4.7 & 2.2 & 0.8 \\
        & ChatGPT (1-shot uncon)  & 7.2 & 4.7 & 1.7 & 0.9 \\
       & ChatGPT (1-shot closest)  & 7.0 & 4.6 & 1.6 & 0.8\\
    \bottomrule
    \end{tabular}
    \caption{Breakdown of WER for ChatGPT error correction results on Transducer and Whisper outputs on LibriSpeech.}
    \label{tab:ct_lb_wer}
\end{table}

We observe that ChatGPT also has a tendency to remove redundant spoken words from the given ASR hypothesis to make the transcription more fluent. With \textit{1-shot closest}, we search from the given N-best list for the final output and therefore the introduced deletion errors can be reduced.   There are more deletions in the Whisper baseline output compared to the Transducer's. When we apply ChatGPT zero-shot unconstrained error correction substitution errors are reduced while more deletions are introduced. Again \textit{1-shot closest}
reduces substitution errors without increasing deletion errors.

To further study the possible reasons why ChatGPT is less effective on Whisper outputs in some cases, we analyse the N-best list of both ASR models, as shown in Table~\ref{tab:cross_wer_break}. 
When computing the statistics, punctuation and special symbols are removed from the ASR hypotheses, leaving only English characters and numbers, to focus on meaningful content.
The \textit{Uniq} metric refers to the average number of unique hypotheses within one N-best list in the test set. For Transducer outputs, the result is close to 5 which is the size of the N-best list, however, there are more repeated entries in Whisper outputs. This is due to the fact that Whisper learns to generate sentences with inverse text normalisation (ITN) to improve the readability, i.e. capitalisation added, punctuation included, and disfluencies removed. As a result, in many cases multiple hypotheses in an N-best list only differ in format, not in actual content. Nevertheless, the diversity of the N-best list is important for our proposed methods to perform well.

\begin{table}[!htbp]
    \centering
    \begin{tabular}{l|l|c|c|c|c|c}
    \toprule
    \multirow{2}*{Data} & \multirow{2}*{Model} & \multirow{2}*{Uniq} & \multicolumn{4}{c}{Cross WER} \\
       &  &  & All & Sub & Del & Ins \\
      \midrule
        \multirow{2}{*}{LB} & Transducer & 4.9 & 9.1 & 7.1 & 1.0 & 1.0 \\
        & \multirow{1}{*}{Whisper} & 3.0 & 12.9 & 7.5 & 2.7 & 2.7 \\
        \midrule
        \multirow{2}{*}{TED} & Transducer & 5.0 & 7.4 & 5.4 & 1.0 & 1.0 \\
         & \multirow{1}{*}{Whisper} & 2.6 & 9.9 & 3.9 & 3.0 & 3.0 \\
         \midrule
        \multirow{2}{*}{Artie} & Transducer & 4.8 & 19.9 & 15.3 & 2.3 & 2.3 \\
         & \multirow{1}{*}{Whisper} & 2.9 & 21.1 & 14.5 & 3.3 & 3.3 \\
        \bottomrule
    \end{tabular}
    \caption{Statistics of the ASR 5-best lists generated by the Conformer-Transducer and the Whisper model on LibriSpeech (LB), TED-LIUM3 (TED) and Artie test sets.}
    \label{tab:cross_wer_break}
\end{table}

Another observation is that for Whisper, even when the hypotheses in the N-best list are diverse, the difference may come from one hypothesis omitting or inserting some irrelevant words in the output.
This is illustrated with the Cross WER metric in Table \ref{tab:cross_wer_break}. Here, we keep all the unique hypotheses in an N-best list. Then for each pair of hypotheses in the remaining list, we calculate the WER result against each other and sum the result on the entire set. This metric can help us measure the difference between hypotheses within one N-best list. The results show that the deletion and insertion rates of Whisper on Cross WER are much higher than the Transducer model, especially on TED-LIUM3. This suggests that Whisper may fail to faithfully transcribe the utterance in all N-best hypotheses, resulting in sentences with varying lengths. ChatGPT tends to choose more coherent ones, leading to many deletion errors in the output, which might explain why ChatGPT underperforms on the TED-LIUM3 data for Whisper.

In Table \ref{tab:case_analysis}, we conduct case analysis for an error correction example from the test set of TED-LIUM3. As the table shows, for the Transducer ASR model, all the hypotheses are of similar length containing all the information from the utterance, and the \textit{Uniq} metric is 5. ChatGPT helps to correct ``blue'' into ``blew'' utilising the given N-best list and world knowledge. Meanwhile, for Whisper 5-best hypotheses, the \textit{Uniq} metric is only 3 due to the repetition problem. In addition, disfluencies in the utterance (``that'', ``you know'') and the non-existent word (``and'') are incorrectly removed or introduced in the output, resulting in more deletions and insertions in Cross WER. The produced N-best list is hence less informative and misleads ChatGPT into the wrong output.

\subsection{Ablation and Analysis}
\label{sec:exp-ablation}
The results on the standard test sets indicate that ChatGPT is effective at detecting errors in the given ASR hypotheses and generating the corrected transcription, especially for out-of-domain scenarios. To further study where the performance gain comes from, we built a ROVER-based system~\cite{fiscus1997post} to align and combine the hypotheses in an N-best list with weighted voting, but it leads to worse results compared to the ASR baseline. It suggests that ChatGPT leverages its implicitly learned world knowledge to generate the corrected ASR transcription based on the given input information, instead of performing a simple voting process on the N-best list.

\begin{table}[htbp!]
    \centering
    \begin{tabular}{l|l|ccc}
        \toprule
        Method & Input & LB & TED & Artie \\
        \midrule
        ASR baseline & - & 6.90 & 13.53 & 23.67\\
        \midrule
        \multirow{4}*{0-shot uncon} & 1-best & 8.25  & 11.95 & 21.19 \\
        & 3-best & 7.01 & 11.31 & 18.84\\
        & 5-best & \textbf{6.64} & 11.35 & 18.73\\
        & 10-best & 6.69 & \textbf{11.29} & \textbf{18.72} \\
        \midrule
        \multirow{3}*{0-shot select} & 3-best & \textbf{6.50} & 12.73 & 22.29\\
        & 5-best & 6.52 & 12.61 & 21.88\\
        & 10-best & 6.58 & \bf 12.56 & \bf 21.47\\
        \bottomrule
    \end{tabular}
    \caption{Ablation of the size of the input N-best list using  the Conformer-Transducer outputs on three test sets.}
    \label{tab:ablation}
\end{table}

In table \ref{tab:ablation}, we take the Conformer-Transducer model as an example to study the influence of the size of the N-best list. Results show that using a large number of N is important for ChatGPT to perform well with the \textit{zero-shot uncon} prompt, especially for the in-domain setting. In the extreme case of only using the top one ASR hypothesis as input, ChatGPT makes many unnecessary changes to the input to make the sentence more ``reasonable'' due to lack of information. With the increased N-best list size, it learns to compare the differences between the hypotheses and correct when the sentences disagree with each other. For the selective approach, the size of the N-best list matters less as ChatGPT performs choice selection rather than generating the entire corrected hypothesis.



\begin{figure}[!htbp]
    \centering
    \includegraphics[width=0.8\linewidth]{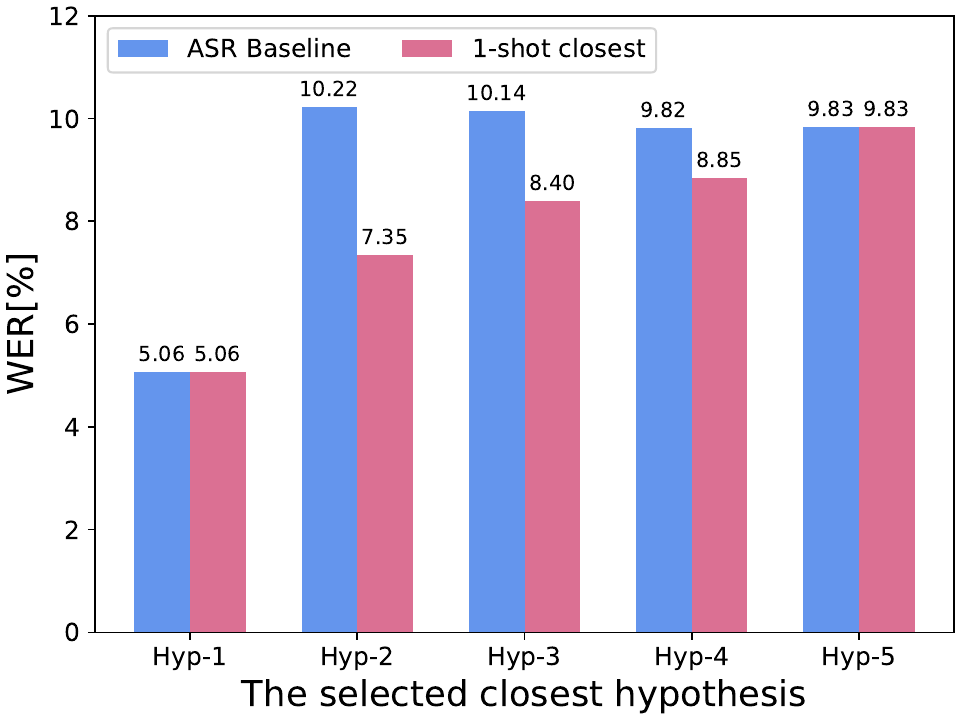}
    \vspace{-0.3cm}
    \caption{Baseline WER of Transducer and error correction results with \textit{1-shot closest}. The LibriSpeech test set is split into 5 parts according to the number of the closest hypothesis.}
    \label{fig:barchart}
\end{figure}

With \textit{1-shot closest} mapping, we select the ASR hypothesis within the N-best list that is most similar to the ChatGPT output. Thus, for each utterance, the selected hypothesis falls in the range of Hyp-1 to Hyp-5, and we divide the LibriSpeech test set into 5 splits accordingly. The proportions of each subset are 67\%, 14\%, 8\%, 5\%, and 6\%. In Figure \ref{fig:barchart}, we show the WER results of the ASR baseline and after error correction for each subset. When Hyp-1 is selected, WER remains the same as the ASR baseline. 
The figure shows that the largest performance gain is achieved when the second best hypothesis from the N-best list is picked. The improvement steadily declines when the number of the selected hypothesis increases until the WER after error correction matches the original ASR baseline when Hyp-5 is selected.



\section{Conclusions}
In this paper we investigate the use of a powerful generative large language model, ChatGPT, to perform ASR error correction in zero-shot and 1-shot settings. Using only information from the ASR N-best list, the system is able to correct errors either by selecting one of the N-best, constrained correction, or in an unconstrained fashion. The proposed methods show gains over the ASR baseline in general for two state-of-the-art system architectures: Transducer and AED. On the Transducer outputs, ChatGPT performs similarly or better than a bespoke ASR error correction system based on the T5 foundation model. For the AED system, it underperforms T5 due to inherent problems with the AED N-best lists.

\bibliographystyle{IEEEbib}
\bibliography{refs}

\end{document}